\documentclass[11pt]{article}

\usepackage{array}
\PassOptionsToPackage{table}{xcolor}
\usepackage[preprint]{acl}
\usepackage{times}
\usepackage{latexsym}
\usepackage[T1]{fontenc}
\usepackage[utf8]{inputenc}
\usepackage{microtype}
\usepackage{graphicx}
\usepackage{amsmath,amssymb,amsfonts}
\usepackage{booktabs}
\usepackage{multirow}
\usepackage{float}
\usepackage{dblfloatfix}
\usepackage{makecell}
\graphicspath{{figures/}}
\newcommand{\eg}{e.g.}

\newcommand{\tblsec}[1]{\multicolumn{8}{>{\columncolor{gray!15}}l}{\rule{0pt}{2.1ex}\textbf{#1}}}
\newcommand{\tcell}[1]{\makecell[c]{#1}}
\newcommand{\lcell}[1]{\makecell[l]{#1}}

\begin{document}

\title{Beyond Scene Priors: Fine-Grained Traffic Scene Reasoning with Benchmarking and Query-Guided Small-Object Focus}

\author{%
Waikit Xiu$^{1,\dagger}$ \qquad Qiang Lu$^{2,\dagger}$ \qquad
\mbox{Zian Wang}$^{1}$ \qquad \mbox{Xinjie Yang}$^{1}$ \\
\bfseries\mbox{Zhiwei Chen}$^{3}$ \qquad Chen Sun$^{1,*}$ \qquad Xiying Li$^{2}$ \\
\normalfont
$^{1}$The University of Hong Kong \qquad $^{2}$Sun Yat-Sen University \\
$^{3}$The Hong Kong University of Science and Technology (Guangzhou) \\
$^{\dagger}$Equal contribution \qquad $^{*}$Corresponding author
}

\maketitle

\begin{abstract}

In safety-critical traffic scenarios, answering complex questions relies on minute, localized visual cues. However, standard Multimodal Large Language Models (MLLMs) tend to over-attend to backgrounds, overwhelming crucial small objects during visual-language alignment, a failure mode we term 'critical evidence dilution.' Furthermore, existing visual question answering (VQA) datasets rarely expose this flaw, as they lack large-scale, distractor-heavy evaluations that require pinpointing local evidence.
To bridge this evaluation and architecture gap, we introduce the Fine-Grained Traffic Reasoning Benchmark (FGTR-Bench) and the Text-Guided Small-Object Reasoning MLLM (TSR-MLLM). FGTR-Bench comprises 40,236 single-image Multiple-Choice Questions (MCQs) created via multi-agent generation, consistency checks, and expert audits, alongside a disjoint 4,947-sample blind test split.
To resolve evidence dilution, TSR-MLLM, built on Qwen3-VL-4B, uses a query-conditioned Text-Guided Small-Object Focus (TG-SOF) map. Applied once at the decoder boundary, the map adds sparse Top-$K$ gated residuals to the most question-relevant vision slots while leaving text tokens unchanged. Together with lightweight decoder adaptation, TSR-MLLM preserves single-pass inference without external detectors or image re-encoding.
Under matched settings, TSR-MLLM outperforms the strongest 4B baseline by 2.1 points on FGTR-Bench (74.1\% overall), with larger gains on evidence-local tracks. Furthermore, it remains competitive on DriveQA-V (CARLA Signs) under greedy decoding without task-specific fine-tuning.

\end{abstract}

\begin{figure}[!t]
    \centering
    \includegraphics[width=\columnwidth,trim=1cm 1cm 1cm 0.5cm,clip]{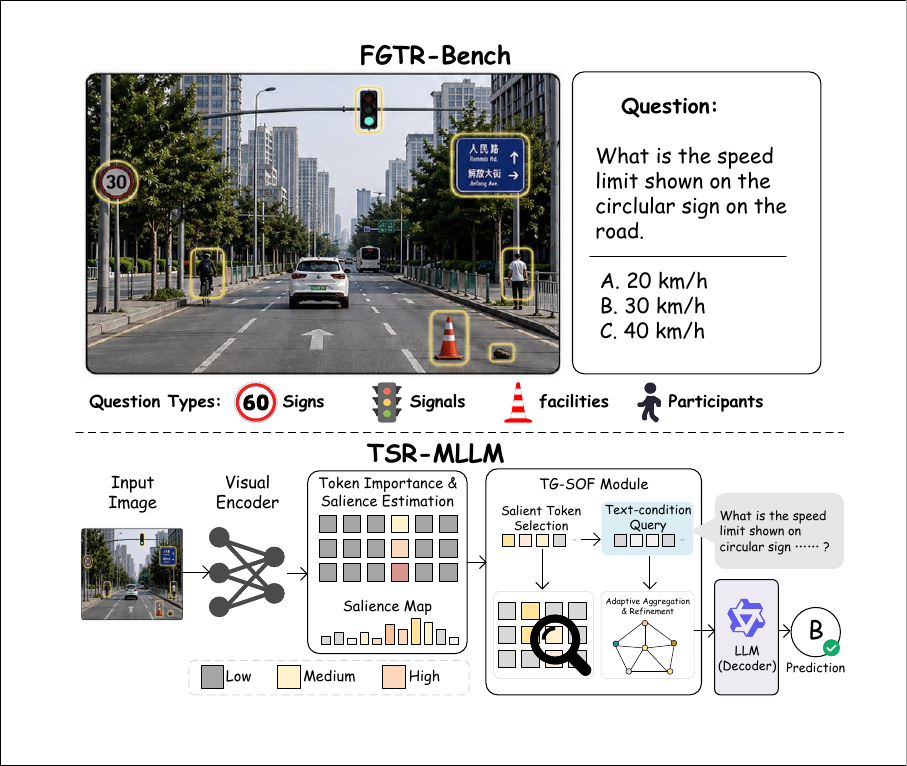}
    \caption{\textbf{Teaser.} \textbf{Top:} FGTR-Bench---single-image traffic MCQs whose answers depend on small, localized evidence.
    \textbf{Bottom:} TSR-MLLM adds TG-SOF before the decoder: query-guided salience, sparse Top-$K$ vision updates, then one forward pass through frozen Qwen3-VL-4B (decoder LoRA optional at train time).}
    \label{fig:teaser}
\end{figure}

\section{Introduction}
Multimodal Large Language Models (MLLMs) have achieved strong progress in general visual question answering (VQA) and cross-modal reasoning \cite{llava,blip2,instructblip,qwenvl,qwen25vl,qwen3vl,internvl,internvl25,llavaov}. However, they still exhibit a persistent bottleneck in fine-grained, text-driven visual reasoning: when answers depend on low-occupancy, local, and sparse cues, the relevant evidence is easily overwhelmed by large background context during full-image token fusion. Consequently, models tend to produce superficially plausible hallucinated responses based on language priors or global scene context rather than decisive local evidence  \cite{shallowfocus,lookshallower,leomini,zoomeye}. We refer to this failure mode as \textit{critical evidence dilution}.  The core issue is not a failure of visual perception, but rather that during vision-language alignment, models fail to effectively retrieve and exploit critical fine-grained visual tokens conditioned on the language query.

Traffic environments provide a uniquely demanding testbed for this cross-modal reasoning challenge. High-stakes queries in these scenarios often hinge on subtle visual entities, such as distant signs, ambiguous road markings, or temporary barriers. Yet, existing traffic datasets predominantly formulate this as a computer vision detection task \cite{tt100k,cctsdb,bdd100k}, while general VQA benchmarks, despite their broad open-ended coverage \cite{vqav2,gqa,aokvqa}, lack the fine-grained semantic granularity and strict evidence-locality constraints required for complex spatial reasoning in traffic safety contexts \cite{drivelm,driveqa,trafficmllm,pope,docvqa}. Consequently, current VQA evaluations leave a critical gap: they cannot distinguish whether an MLLM is conducting rigorous logical reasoning over localized visual evidence or merely matching global scene statistics with language priors. This evaluation gap obscures the true multimodal reasoning capabilities of current models.

To address these gaps, we release FGTR-Bench, an evidence-grounded benchmark specifically designed to evaluate fine-grained traffic reasoning in MLLMs. Comprising 45,183 high-quality question-answering pairs derived from both public and in-house data, FGTR-Bench unifies challenging semantic tasks spanning holistic sign interpretation, daytime signal reading, nighttime signal reading, roadside micro-hazards, and participant micro-risk. To ensure high-fidelity text-vision alignment, we employ a multi-agent LLM pipeline for QA generation, coupled with structural consistency checks and rigorous manual auditing. This guarantees strict causal alignment between the text queries, multiple-choice options, and the localized visual evidence.

Building upon this benchmark, we propose Text-Guided Small-Object Reasoning MLLM (TSR-MLLM). To counteract evidence dilution, we introduce a Text-Guided Small-Object Focus (TG-SOF) module prior to text decoding. Instead of relying on external object detectors, TG-SOF utilizes the linguistic query to compute cross-modal attention, applying controlled residual updates to a dynamically selected subset of highly relevant visual tokens. Furthermore, we introduce an evidence-chain supervision mechanism during the instruction-tuning phase. By integrating a bounding-box alignment loss, we inject explicit spatial-grounding signals into the language modeling objective. This forces the model to anchor its text generation on the correct visual regions, drastically improving both the fidelity of the fine-grained reasoning and the interpretability of the generated text.

Our contributions are summarized as follows:
\begin{itemize}
    \item We release FGTR-Bench, a comprehensive vision-language benchmark designed to systematically evaluate MLLMs' fine-grained reasoning and evidence-grounding capabilities in high-risk, small-object-driven scenarios.
    \item We propose TSR-MLLM, a new framework with query-guided TG-SOF that improves local critical-evidence modeling without increasing inference pipeline complexity.
    \item We design a spatially-grounded auxiliary supervision strategy for MLLM instruction tuning. By aligning text generation with explicit evidence chains, we significantly improve the model's reasoning stability and interpretability.
\end{itemize}

\section{Related Work}
\noindent\textbf{Traffic perception and VQA.}
Traffic small-object research is still largely detection-centric, with representative lines including multi-scale fusion (FPN~\cite{fpn}), two-stage refinement (Mask R-CNN~\cite{maskrcnn}), and stronger backbones/adapters for robustness under long range, low resolution, occlusion, and nighttime conditions~\cite{swin,vit_adapter,tinyperson,visdrone}. In traffic datasets such as TT100K, CCTSDB, and BDD100K~\cite{tt100k,cctsdb,bdd100k}, this typically remains a closed-set objective (detect/classify predefined categories). In parallel, traffic VQA has moved from explicit QA toward language-grounded driving understanding~\cite{drivelm,driveqa,trafficmllm,talk2drive}, but most benchmarks still emphasize scene-level reasoning or local attribute querying~\cite{vqav2,gqa,docvqa}, often on perception-first driving data~\cite{nuscenes,waymo,uniad,genad}. As a result, decision-oriented QA triggered by sparse local evidence is still weakly covered, especially under unified distractor-heavy evaluation.

\begin{figure*}[!t]
    \centering
    \includegraphics[width=0.9\textwidth,trim=1.2cm 0.5cm 0.5cm 0.5cm,clip]{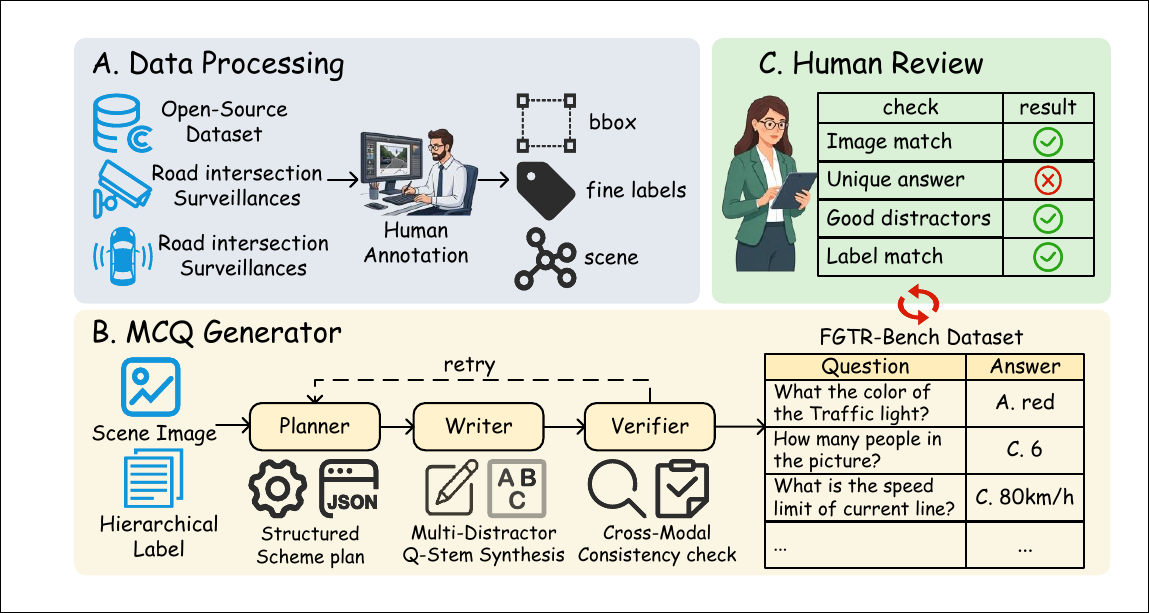}
    \caption{TSR-Gen data pipeline for FGTR-Bench. A, Data processing: open-source and surveillance imagery is human-annotated into bbox, fine labels, and a scene graph. B, MCQ generator: a scene image and hierarchical labels feed a Planner--Writer--Verifier loop with retry, producing FGTR-Bench MCQs. C, Human review: reviewers check image match, unique answer, good distractors, and label match before release.}
    \label{fig:pipeline}
\end{figure*}

\noindent\textbf{MLLMs and fine-grained perception.} 
MLLMs move VQA from closed-set prediction to open-ended generation, as seen in LLaVA/BLIP-style systems and recent Qwen/InternVL/DeepSeek families~\cite{llava,blip2,qwenvl,qwen25vl,internvl,llavaov,emu2,deepseekv3,deepseekr1}. This trend is supported by large-scale vision-language pretraining and instruction data pipelines~\cite{vit,visualwebinstruct}, and common language backbones~\cite{llama,vicuna}. In fine-grained traffic reasoning, decisive cues are sparse and easily diluted by global fusion. Existing remedies include zooming and multi-pass processing~\cite{zoomeye,zwz,zwzr2i}, external detector/OCR signals~\cite{knowlook}, and training-free attention steering~\cite{leomini,shallowfocus,lookshallower}; related reasoning-control lines~\cite{seeingbelieving,realign,cacheofthought} point to similar bottlenecks. Closely related training-free selectors further retain query-relevant visual tokens under a fixed budget~\cite{script,flashvlm,rediprune}, primarily by dropping redundant pre-decode tokens for efficiency. TG-SOF instead targets critical evidence dilution in traffic MCQ: it keeps the full vision grid and applies a learned, sparse query-conditioned Top-$K$ residual on fused decoder inputs in one pass, without shortening the visual sequence, external detectors, or re-encoding.

\section{FGTR-Bench: A Fine-Grained Traffic Reasoning Benchmark}

FGTR-Bench unifies small-object-driven traffic reasoning under a single evaluation protocol, designed around visual evidence locality rather than coarse task categories. Specifically, we couple each question to verifiable local evidence and enforce same-scene hard negatives during option construction.This ensures we can report disaggregated accuracy by decision-critical regimes, allowing performance gains to be attributed to true evidence utilization rather than reliance on scene priors (Appendices~\ref{app:dataset},~\ref{app:eval}).

\subsection{Task Definition and Taxonomy}
We define fine-grained traffic reasoning as an image-grounded discrete decision problem. Each instance is represented as a tuple \(z = (\mathbf{x}, q, \{c_\ell\}_{\ell=1}^{4}, y^*)\), where \(\mathbf{x}\) is an RGB frame, \(q\) is the question stem, $\{c_\ell\}_{\ell=1}^{4}$ are four candidate options, and \(y^*\) is the unique correct choice. After multimodal fusion, a model with parameters \(\theta\) defines a probability distribution\(p_\theta(y \mid \mathbf{x}, q, c_{1:4})\). At inference, we decode a single prediction \(\hat{y}\) and evaluate accuracy over held-out samples.

To quantitatively evaluate evidence utilization, each sample is assigned to one of five tracks: holistic sign, daytime signal, nighttime signal, roadside micro, and participant micro. This taxonomy organizes the benchmark around two complementary reasoning regimes: semantic control interpretation (e.g., sign meaning, lane-linked disambiguation, multi-sign composition, day/night signal reading) and ground-level micro-risk understanding (e.g., distant participants, cones, debris). We report both per-track and overall accuracy to expose the specific domains driving model performance.

\subsection{Data Sources and Pre-processing}
FGTR-Bench is constructed from both public datasets and in-house collection to cover critical safety scenarios absent from any single source. We use TT100K~\cite{tt100k} and LISA~\cite{lisa} with extensive restructuring, including finer-grained relabeling for TT100K and strict daytime/nighttime partitioning for LISA. We further incorporate collected roadside and vehicle-mounted real-world imagery to supplement participant interactions, facility-level details, and road-surface micro hazards that are underrepresented in existing corpora.

All images are manually annotated with high-precision bounding boxes and fine-grained semantic tags for small, safety-relevant entities. This produces a structured annotation layer that serves as a hard constraint for downstream QA construction, ensuring that generated reasoning instances remain strictly grounded in verifiable visual evidence.

\subsection{TSR-Gen: QA Construction and Alignment}
We develop TSR-Gen (Figure~\ref{fig:pipeline}) to convert structured perception annotations into reasoning-oriented QA while preserving strict image-question alignment. For each image, we first construct a scene topology graph \(\mathcal{G}=(\mathcal{V}_{\mathrm{ent}},\mathcal{E})\), where nodes store boxes, coarse/fine labels, and optional lane/phase tags, and edges encode spatial or relational constraints from manual or high-confidence automatic labels.

TSR-Gen then packages \(\mathcal{G}\) and the raw frame into a unified ground-truth representation \(\mathcal{B}\). An LLM-driven multi-agent pipeline maps  \(\mathcal{B}\) to the final QA instances: a Planner selects template type and target evidence entities, a Writer generates the question and options, and a Verifier enforces consistency among labels, boxes, and textual statements before export. Category-level consistency filters, followed by expert triage, further minimize hallucinated visual references. Finally, a post-filter manual audit protocol is applied prior to release. The released development corpus contains 40{,}236 instances (34{,}749 train / 5{,}487 validation) with disjoint image paths across splits; all main-paper benchmark numbers are reported on an additional disjoint blind test set of 4{,}947 samples (Appendix~\ref{app:data_stats}).

\section{TSR-MLLM: Text-Guided Small-Object Reasoner}
\label{sec:tsr_mllm}

\subsection{Overall Architecture}
Let $(\mathbf{x},q)$ denote an image--query pair.
For prompt length $L$, visual token indices are $\mathcal{V}\subseteq\{1,\ldots,L\}$, and text indices $\mathcal{T}$ are the remaining positions in $\{1,\ldots,L\}$ that are not in $\mathcal{V}$.
TSR-MLLM follows the Qwen3-VL multimodal layout: a frozen vision tower and text tokenizer populate fused decoder inputs $\mathbf{H}^{(0)}\in\mathbb{R}^{L\times d}$ with text rows at $\mathcal{T}$ and vision rows at $\mathcal{V}$.
We do not add auxiliary detectors, change the fusion topology, re-encode the image, or insert tool-augmented perception loops.

TG-SOF is the only structural insertion: a thin map $\mathcal{F}_{\phi}$ applied once after assembly and before the causal decoder stack.
The map reads both $\mathbf{H}^{(0)}$ and the query text that conditions the task, and returns refined decoder inputs:
\begin{equation}
\label{eq:tg_sof_map}
\mathbf{H}^{(1)} = \mathcal{F}_{\phi}(\mathbf{H}^{(0)}, q).
\end{equation}
Text rows are fixed, $\mathbf{H}^{(1)}_{t,:}=\mathbf{H}^{(0)}_{t,:}$ for all $t\in\mathcal{T}$.
Only a Top-$K$ sparse subset of vision rows in $\mathcal{V}$ receives bounded residuals initialized near zero so optimization begins near the pretrained interface.
The subsequent forward pass is unchanged from the base model: $\mathcal{G}_{\theta}$ consumes $\mathbf{H}^{(1)}$, followed by vocabulary projection, and FGTR-Bench multiple-choice answers are read off by argmax over choice logits.

\begin{figure*}[!t]
    \centering
    \includegraphics[width=\textwidth,trim=1.2cm 0.5cm 2.4cm 1.2cm,clip]{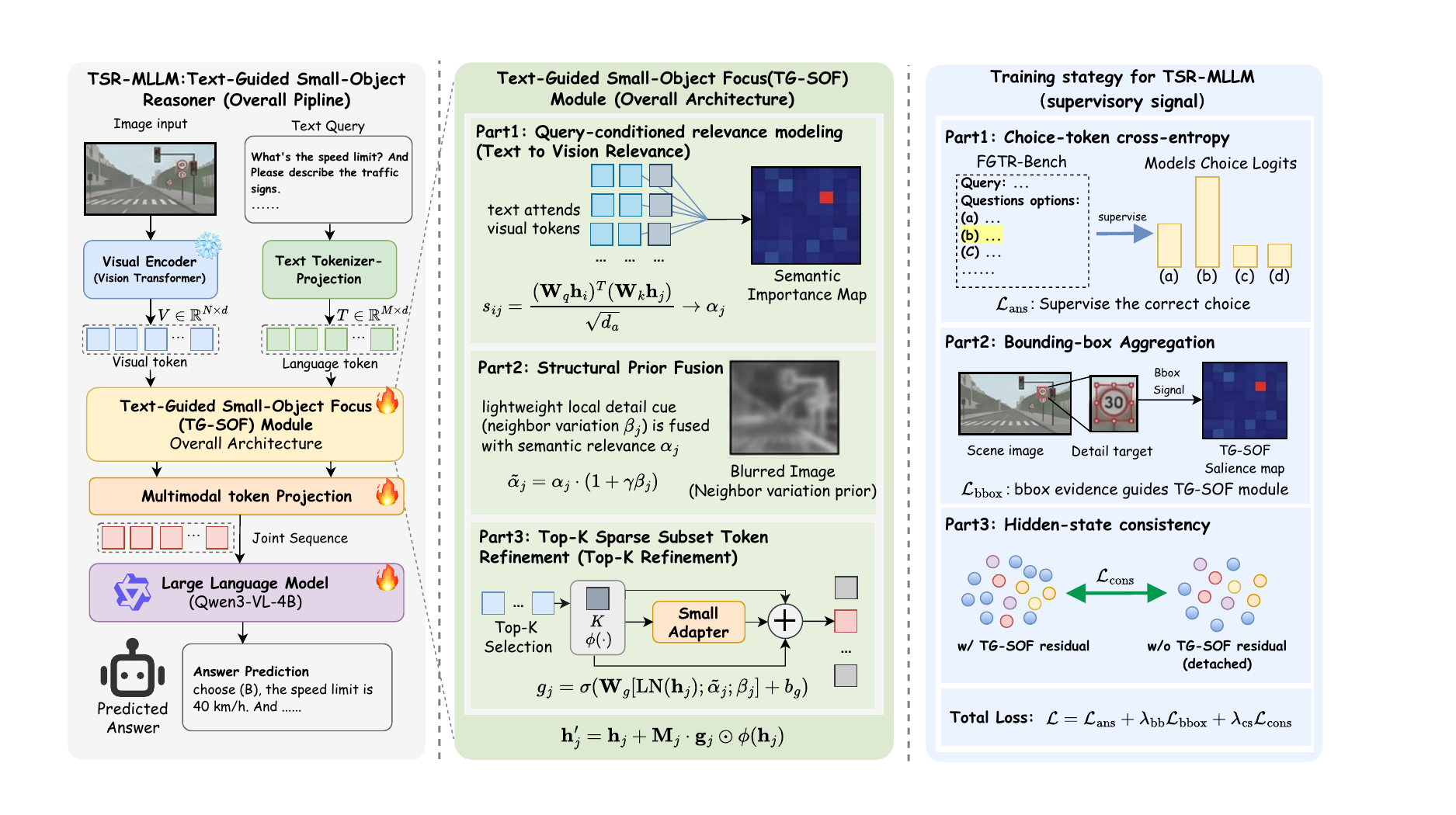}
    \caption{Overview of TSR-MLLM.
    \textbf{Left:} The end-to-end inference pipeline. An image and text query are encoded into multimodal tokens, refined by TG-SOF before decoding, and the model predicts a multiple-choice answer in a single forward pass without external detectors or re-encoding.
    \textbf{Middle:} The TG-SOF module. It computes query--vision salience, sharpens it with a local detail prior, and applies sparse Top-$K$ gated residuals to the most question-relevant vision tokens.
    \textbf{Right:} The training strategy on FGTR-Bench. Three objectives supervise answer selection, bbox-guided evidence focus, and hidden-state consistency during fine-tuning.}
    \label{fig:tsr_mllm_framework}
\end{figure*}

\subsection{Text-Guided Small-Object Focus}
\label{sec:tg_sof}
Fine-grained traffic questions often hinge on a few vision tokens, yet those tokens are easily drowned out by large background fields before deep decoder self-attention.
TG-SOF operates at this boundary in two linked stages: query-conditioned salience scoring with a local detail sharpen, followed by sparse Top-$K$ gated residual updates on vision rows only.

\paragraph{Query--vision salience with a local sharpen.}
Let $\mathcal{T}_{\mathrm{qv}}\subseteq\mathcal{T}$ index query-visible task text---the MCQ stem and option block after templating---and $\mathbf{h}^{(0)}_t:=\mathbf{H}^{(0)}_{t,:}$.
For $i\in\mathcal{T}_{\mathrm{qv}}$ and $j\in\mathcal{V}$, bilinear query--vision scores are are calculated as follows:
\begin{equation}
\label{eq:score_ij}
s_{ij}=\frac{(\mathbf{W}_q\mathbf{h}^{(0)}_i)^\top(\mathbf{W}_k\mathbf{h}^{(0)}_j)}{\sqrt{d_a}}.
\end{equation}
Averaging the implied query-to-vision softmax masses over $i$ yields salience $\alpha_j$:
\begin{equation}
\label{eq:salience}
\alpha_j = \frac{1}{|\mathcal{T}_{\mathrm{qv}}|}\sum_{i\in\mathcal{T}_{\mathrm{qv}}}\frac{\exp(s_{ij})}{\sum_{k\in\mathcal{V}}\exp(s_{ik})}.
\end{equation}
Pure $\alpha_j$ can track broad texture that correlates with language; we therefore sharpen rankings with
\begin{equation}
\label{eq:sharpen}
\tilde{\alpha}_j=\alpha_j(1+\gamma\beta_j),
\end{equation}
for $\gamma\ge 0$, where $\beta_j$ measures short-range contrast along the raster-ordered vision sequence so that sharp micro-structure can enter the Top-$K$ set.

\paragraph{Top-$K$ gated residual.}
We select the Top-$K$ indices under $\tilde{\alpha}_j$ with $K=\max\{1,\lfloor \rho|\mathcal{V}|\rfloor\}$ and form a hard mask $m_j\in\{0,1\}$ on the corresponding vision rows.
Only masked slots receive a gated residual: a scalar $\lambda_j\in(0,1)$ scales the output of a narrow LayerNorm MLP $\psi_{\phi}$, and $\lambda_j$ is computed from pooled query states together with $\tilde{\alpha}_j$, $\beta_j$, and $\mathbf{h}^{(0)}_j$.
For vision indices $j\in\mathcal{V}$,
\begin{equation}
\label{eq:tg_sof_update}
\mathbf{H}^{(1)}_{j,:} = \mathbf{H}^{(0)}_{j,:} + m_j\,\lambda_j\,\psi_{\phi}\!\bigl(\mathrm{LN}(\mathbf{h}^{(0)}_j)\bigr).
\end{equation}
Text rows are unchanged, with $\mathbf{H}^{(1)}_{t,:}=\mathbf{H}^{(0)}_{t,:}$ for all $t\in\mathcal{T}$.
We initialize $\psi_{\phi}$ and the gate path near zero so $\mathcal{F}_{\phi}$ starts close to identity on vision rows.

\begin{table*}[!t]
    \centering
    \small
    \setlength{\tabcolsep}{4pt}
    \renewcommand{\arraystretch}{1.05}
    \begin{tabular*}{\textwidth}{@{\extracolsep{\fill}}lccccccc}
    \toprule
    \tcell{Model} & \tcell{Params}
    & \tcell{Holistic\\Sign $\uparrow$}
    & \tcell{Daytime\\Signal $\uparrow$}
    & \tcell{Nighttime\\Signal $\uparrow$}
    & \tcell{Roadside\\Micro $\uparrow$}
    & \tcell{Participant\\Micro $\uparrow$}
    & \tcell{Overall\\Acc. $\uparrow$} \\
    \midrule\noalign{\vskip-2pt}
    \tblsec{w/o FGTR-Bench training} \\
    \lcell{LLaVA-1.5~\cite{llava15}} & \tcell{7B} & \tcell{41.9} & \tcell{77.5} & \tcell{29.3} & \tcell{14.4} & \tcell{42.0} & \tcell{41.0} \\
    \lcell{LLaVA-OneVision~\cite{llavaov}} & \tcell{7B} & \tcell{83.0} & \tcell{80.8} & \tcell{75.6} & \tcell{27.5} & \tcell{53.3} & \tcell{65.4} \\
    \lcell{Qwen2.5-VL-3B~\cite{qwen25vl}} & \tcell{3B} & \tcell{82.0} & \tcell{73.2} & \tcell{64.7} & \tcell{29.1} & \tcell{36.6} & \tcell{58.5} \\
    \lcell{Qwen2.5-VL-7B~\cite{qwen25vl}} & \tcell{7B} & \tcell{86.7} & \tcell{97.9} & \tcell{77.7} & \tcell{34.0} & \tcell{51.3} & \tcell{69.9} \\
    \lcell{Qwen3-VL-4B~\cite{qwen3vl}} & \tcell{4B} & \tcell{86.8} & \tcell{96.3} & \tcell{\textbf{83.2}} & \tcell{28.9} & \tcell{52.4} & \tcell{70.1} \\
    \midrule\noalign{\vskip-2pt}
    \tblsec{w/ FGTR-Bench training} \\
    \lcell{LLaVA-1.5~\cite{llava15}} & \tcell{7B} & \tcell{73.0} & \tcell{79.7} & \tcell{48.2} & \tcell{14.2} & \tcell{30.4} & \tcell{50.5} \\
    \lcell{LLaVA-OneVision~\cite{llavaov}} & \tcell{7B} & \tcell{86.8} & \tcell{87.9} & \tcell{78.4} & \tcell{29.6} & \tcell{\textbf{54.1}} & \tcell{68.5} \\
    \lcell{Qwen2.5-VL-3B~\cite{qwen25vl}} & \tcell{3B} & \tcell{85.4} & \tcell{79.1} & \tcell{67.9} & \tcell{30.7} & \tcell{39.4} & \tcell{61.8} \\
    \lcell{Qwen2.5-VL-7B~\cite{qwen25vl}} & \tcell{7B} & \tcell{90.0} & \tcell{98.2} & \tcell{80.0} & \tcell{\textbf{35.1}} & \tcell{52.8} & \tcell{71.8} \\
    \lcell{Qwen3-VL-4B~\cite{qwen3vl}} & \tcell{4B} & \tcell{94.1} & \tcell{97.4} & \tcell{83.1} & \tcell{28.4} & \tcell{51.4} & \tcell{72.0} \\
    \midrule\noalign{\vskip-2pt}
    \tblsec{Ours} \\
    \lcell{\textbf{TSR-MLLM (ours)}} & \tcell{\textbf{4B}} & \tcell{\textbf{97.4}} & \tcell{\textbf{99.3}} & \tcell{\textbf{86.9}} & \tcell{29.4} & \tcell{52.3} & \tcell{\textbf{74.1}} \\
    \bottomrule
    \end{tabular*}
    \caption{FGTR-Bench accuracy (\%) across subtasks and overall performance. Rows represent models without FGTR-Bench training, FGTR-Bench fine-tuned models and TSR-MLLM. All numbers are evaluated on the blind test set (1{,}447 holistic, 750 daytime, 750 nighttime, 800 roadside, 1{,}200 participant; 4{,}947 overall). Models are trained on the training split only; the test set is never used for model selection.}
    \label{tab:main_results}
  \end{table*}

\subsection{Training Strategy and Optimization}
\label{sec:train_opt}
We fine-tune TSR-MLLM on FGTR-Bench with the Qwen3-VL backbone frozen and lightweight modules trained on top.
The overall objective is
\begin{equation}
\label{eq:total_loss}
\mathcal{L} = \mathcal{L}_{\text{ans}} + \lambda_{\text{bb}}(t)\,\mathcal{L}_{\text{bbox}} + \lambda_{\text{cs}}\,\mathcal{L}_{\text{cons}},
\end{equation}
where $\lambda_{\text{bb}}(t)$ may be linearly warmed from zero.

\textbf{Training strategy.}
Trainable parameters reside in TG-SOF, the vocabulary projection head \texttt{lm\_head}, and decoder LoRA; the vision tower and pretrained decoder blocks remain frozen.
Optimization follows standard causal fine-tuning on FGTR-Bench with matched preprocessing and greedy letter decoding at evaluation.

\textbf{Choice-token cross-entropy ($\mathcal{L}_{\text{ans}}$).}
The primary term is cross-entropy on supervised choice tokens with prompts masked, which supervises the correct multiple-choice letter under image and instruction context.

\textbf{Bounding-box aggregation.}
Each training sample carries axis-aligned boxes; if multiple boxes are listed, we replace them by their pixel union $R$.
For each $j\in\mathcal{V}$, let $a_j=1$ when the patch for vision token $j$ overlaps $R$, and set $\mathcal{B}=\{j\in\mathcal{V}:a_j=1\}$.
If $\mathcal{B}$ is empty, we skip the term; otherwise the teacher is the uniform distribution on evidence tokens, $q^{\text{bbox}}_j=a_j/|\mathcal{B}|$.
TG-SOF already outputs sharpened salience $\tilde{\alpha}_j$.
We reuse it as logits
\begin{equation}
\label{eq:salience_logit}
s_j=\log(\tilde{\alpha}_j+\varepsilon),
\end{equation}
with floor $\varepsilon>0$ and temperature $\tau>0$, and obtain
\begin{equation}
\label{eq:salience_prob}
p_j = \frac{\exp(s_j/\tau)}{\sum_{k\in\mathcal{V}}\exp(s_k/\tau)},\qquad j\in\mathcal{V}.
\end{equation}
The box alignment loss is forward KL on the same scoring pathway that feeds Top-$K$:
\begin{equation}
\label{eq:bbox_loss}
\mathcal{L}_{\text{bbox}} = \sum_{j\in\mathcal{V}} q^{\text{bbox}}_j \log\frac{q^{\text{bbox}}_j}{p_j}.
\end{equation}

\textbf{Hidden-state consistency.}
Let $\mathcal{S}$ denote supervised decoder positions.
With TG-SOF enabled, student hidden states $\hat{\mathbf{z}}_s$ are compared to detached teacher states $\mathbf{z}^{\text{tch}}_s$ from a no-residual forward:
\begin{equation}
\label{eq:cons_loss}
\mathcal{L}_{\text{cons}}=\frac{1}{|\mathcal{S}|}\sum_{s\in\mathcal{S}}\bigl\lVert \hat{\mathbf{z}}_s-\mathbf{z}^{\text{tch}}_s\bigr\rVert_2^2.
\end{equation}
This term limits representation drift from the pretrained interface while still allowing sparse vision updates where $\mathcal{L}_{\text{ans}}$ needs them.

In our reported setup, all three loss terms are active throughout training, with $\lambda_{\text{bb}}(t)$ warmed from zero for stable optimization on the frozen backbone.

\section{Experiments}
\label{sec:experiments}

\begin{table*}[!t]
    \centering
    \small
    \setlength{\tabcolsep}{6pt}
    \begin{tabular}{@{}lccccc c@{}}
    \toprule
    Model & Params & Regulatory$\uparrow$ & Warning$\uparrow$ & Guide$\uparrow$ & Temporary Control$\uparrow$ & Overall$\uparrow$ \\
    \midrule
    Mini-InternVL~\cite{internvl25} & 2B & 64.1 & 55.3 & 65.8 & 45.0 & 59.6 \\
    LLaVA-1.5~\cite{llava15} & 7B & 23.5 & 26.6 & 22.3 & 21.1 & 24.1 \\
    LLaVA-1.6-mistral~\cite{llavaov} & 7B & 42.6 & 43.0 & 52.8 & 37.5 & 43.4 \\
    VILA-1.5~\cite{internvl} & 8B & 25.3 & 23.3 & 27.8 & 21.5 & 24.6 \\
    Traffic-MLLM~\cite{trafficmllm} & 4B & \textbf{75.7} & 74.8 & \textbf{72.1} & 70.6 & 74.5 \\
    TSR-MLLM (ours) & 4B & 74.3 & \textbf{78.2} & 62.7 & \textbf{81.2} & \textbf{74.9} \\
    \bottomrule
    \end{tabular}
    \caption{DriveQA-V (CARLA Signs) accuracy (\%). Bold values denote the optimal performance per column. All models adopt greedy letter decoding and preprocessing consistent with FGTR-Bench. TSR-MLLM receives no fine-tuning on DriveQA. Sample sizes: Regulatory 7985, Warning 5634, Guide 2071, Temporary 1685, Overall 17375.}
    \label{tab:driveqa_signs_results}
\end{table*}

\begin{table}[!t]
    \centering
    \small
    \setlength{\tabcolsep}{3.5pt}
    \renewcommand{\arraystretch}{1.12}
    \begin{tabular}{@{}lcc@{}}
    \toprule
    \lcell{Model} & \tcell{FGTR-Bench\\Overall$\uparrow$} & \tcell{DriveQA-V\\Overall$\uparrow$} \\
    \midrule
    \lcell{Qwen3-VL-4B w/ FGTR-Bench} & \tcell{72.0} & \tcell{71.7} \\
    \lcell{w/ TG-SOF} & \tcell{72.8} & \tcell{73.4} \\
    \lcell{w/ local detail prior} & \tcell{73.3} & \tcell{73.9} \\
    \lcell{w/ bbox focus loss} & \tcell{73.7} & \tcell{74.5} \\
    \lcell{w/ hidden consistency} & \tcell{\textbf{74.1}} & \tcell{\textbf{74.9}} \\
    \bottomrule
    \end{tabular}
    \caption{Cumulative ablation on the FGTR-Bench blind test set (overall accuracy, \%) and DriveQA-V overall (\%). Rows progressively enable modules under the same FGTR-Bench fine-tuning budget.}
    \label{tab:ablation_components}
\end{table}

\subsection{Experimental Setup}

\paragraph{Benchmarks and splits.}
We evaluate TSR-MLLM on two benchmarks: FGTR-Bench for in-domain fine-grained traffic reasoning, and DriveQA-V for out-of-distribution sign QA transfer.
Supervised adaptation uses the 34{,}749-sample FGTR-Bench training split only.
The 5{,}487-sample validation split supports development and qualitative analysis but is not used for the main-table FGTR-Bench numbers.
All accuracy results in Table~\ref{tab:main_results} and Table~\ref{tab:ablation_components} are evaluated on the disjoint blind test set of 4{,}947 samples.

\paragraph{FGTR-Bench tracks and metrics.} We use FGTR-Bench to test whether models exploit decisive local cues rather than scene priors.
It covers five subtasks (holistic traffic signs, daytime/nighttime signals, roadside tiny objects, and road participants), with per-subtask and overall accuracy on the blind test set.

\paragraph{DriveQA-V.}
We use DriveQA-V to verify transfer beyond FGTR-Bench without additional fine-tuning.
Signs are grouped into regulatory, warning, guide, and temporary control families, with per-family and overall accuracy.

\paragraph{Baselines.}
Table~\ref{tab:main_results} lists open MLLMs without FGTR-Bench training, the same model families after FGTR-Bench fine-tuning with optional LoRA~\cite{lora}, and our full TSR-MLLM (Qwen3-VL-4B with TG-SOF, decoder LoRA, and the full training objective in Sec.~\ref{sec:train_opt}).
Families include LLaVA-1.5 and LLaVA-OneVision~\cite{llava15,llavaov}, multiple scales of Qwen2.5-VL and Qwen3-VL~\cite{qwen25vl,qwen3vl}, and broader Qwen-VL and InternVL references~\cite{qwenvl,llava,internvl,internvl25}.
Unless stated otherwise, supervised adaptation for TSR-MLLM and matched baselines uses the FGTR-Bench training split with preprocessing and decoding aligned to evaluation.

\paragraph{Implementation.}
TSR-MLLM freezes Qwen3-VL-4B~\cite{qwen3vl}, inserts TG-SOF before the decoder, and is trained on FGTR-Bench with micro-batch 1 per device, gradient accumulation 8, two data-parallel ranks (effective batch 16), learning rate $5\times10^{-6}$, weight decay 0, and bf16 autocast.
Decoder LoRA uses rank 64, scaling $\alpha{=}128$, dropout 0.05, on all self-attention projections in every decoder layer.
Visual preprocessing uses the $12{,}544$--$12{,}845{,}056$ pixel bounds (resize on, square layout).
Matched baselines follow the same FGTR-Bench schedule and decoding protocol but omit TG-SOF.
Latency and decoding caps follow the same recipe as TSR-MLLM; full TG-SOF, training, and evaluation details are in Appendices~\ref{app:method}--\ref{app:eval}.

\begin{figure*}[!t]
    \centering
    \includegraphics[width=\textwidth,trim=0.4cm 0.4cm 0.4cm 0.4cm,clip]{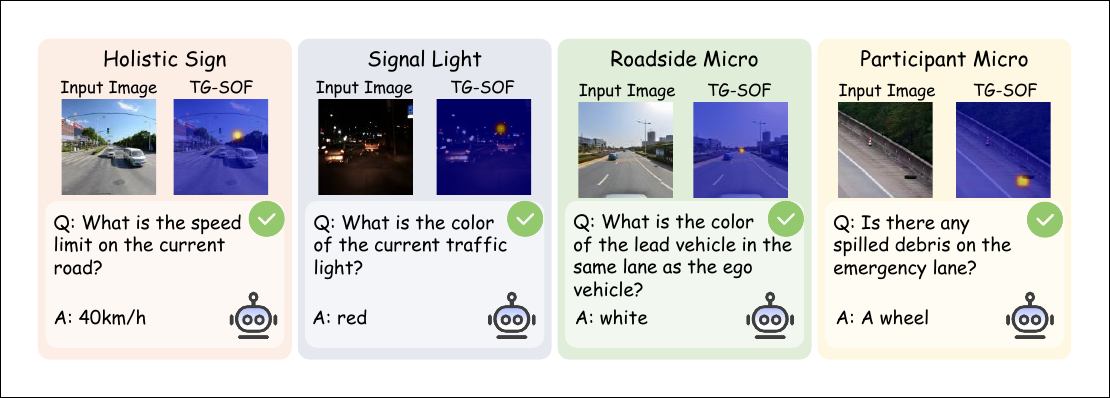}
    \caption{Qualitative TG-SOF visualizations on FGTR-Bench validation items.
    Four columns correspond to \textit{Holistic Sign}, \textit{Signal Light}, \textit{Roadside Micro}, and \textit{Participant Micro}.
    Each panel shows the input frame, the TG-SOF importance overlay, the MCQ stem, and the verified answer ({\checkmark}).
    The highlighted regions illustrate \emph{query--evidence alignment}: TG-SOF concentrates on the visual chain that the question text makes answer-critical; see \S\ref{sec:qualitative}.}
    \label{fig:qualitative}
\end{figure*}

\subsection{Experimental Results}
Table~\ref{tab:main_results} and Table~\ref{tab:driveqa_signs_results} summarize in-domain FGTR-Bench performance and zero-shot DriveQA-V transfer under a matched decoding protocol.

\textbf{FGTR-Bench:} Table~\ref{tab:main_results} reports blind-test results under a unified decoding protocol.
Domain adaptation remains beneficial across model scales, and TSR-MLLM achieves the strongest overall performance among same-scale FGTR-Bench--adapted baselines.
The gains are spread across subtasks rather than concentrated in a single column, with consistent improvements on evidence-local tracks such as \textit{Nighttime Signal}, \textit{Roadside Micro}, and \textit{Participant Micro}.
Although \textit{Roadside Micro} remains the hardest subset in absolute terms, its steady improvement supports the claim that TG-SOF mainly helps where answer-critical evidence is sparse and localized.

\textbf{DriveQA-V:} Table~\ref{tab:driveqa_signs_results} reports out-of-distribution transfer on CARLA Signs under the same greedy decoding setup.
Without any DriveQA fine-tuning, TSR-MLLM reaches the best overall score and outperforms strong open baselines including Traffic-MLLM.
The largest improvements appear on \textit{Warning} and \textit{Temporary Control}, where answers depend on compact sign details in cluttered scenes, while \textit{Regulatory} and \textit{Guide} remain more competitive for specialized baselines.
These results suggest that FGTR-Bench-trained TG-SOF transfers beyond the training benchmark and improves evidence-local sign reasoning under distribution shift.

\subsection{Ablation Studies}

Table~\ref{tab:ablation_components} reports a cumulative ablation under the same FGTR-Bench fine-tuning budget on the blind test set: starting from the FGTR-adapted Qwen3-VL-4B backbone, we progressively enable TG-SOF, the local detail prior, the bounding-box focus loss, and hidden-state consistency, with each row adding one component, and evaluate the same variants on DriveQA-V without DriveQA fine-tuning.
Each module improves both benchmarks monotonically; TG-SOF yields the clearest initial gain by redirecting attention to question-relevant vision tokens, while later components further stabilize fine-grained evidence use, and the full stack reaches the strongest overall results on FGTR-Bench and DriveQA-V.
Appendix~\ref{app:additional} reports five-seed Wilcoxon signed-rank tests confirming that each cumulative step is significant ($p<0.01$).

\subsection{Qualitative Analysis}
\label{sec:qualitative}

Figure~\ref{fig:qualitative} complements Table~\ref{tab:main_results} by testing whether inference-time evidence aligns with the question rather than generic scene context: four validation items spanning holistic signs, signal state, lane participants, and roadside micro-hazards each show the input frame, TG-SOF overlay, MCQ stem, and verified answer ({\checkmark}).
Overlays use sharpened query--vision salience $\tilde{\alpha}_j$ (Sec.~\ref{sec:tg_sof}), projected onto the vision-token grid without gradient attribution; peak regions track the stem's focus---limit signs for speed questions, red signal heads at night, ego-lane lead vehicles, and emergency-lane debris---showing question--evidence unity even for small, low-occupancy targets, consistent with localized gains in Table~\ref{tab:main_results}.
Appendix~\ref{app:additional} adds failure cases under motion blur and low-light glare.

\section{Conclusion}
This work targets \textit{critical evidence dilution} in traffic MCQ reasoning, sparse local cues are easily overwhelmed by background-dominated fusion before decoding.
FGTR-Bench measures evidence use under distractor-heavy multiple choice with verifiable local annotations; TSR-MLLM adds query-guided TG-SOF, a single-pass sparse residual on question-relevant vision tokens at the decoder boundary, without detectors or re-encoding.
Results on FGTR-Bench and DriveQA-V, together with ablations and qualitative maps, support this mechanism rather than a benchmark-specific trick.
Fine-grained traffic reasoning is therefore not only a data or capacity problem, but also an interface problem: models need evidence-local evaluation and explicit token-focus control before decoding.

\section{Limitations}
\label{sec:limitations}
Our experiments are restricted to single-image traffic MCQ tasks. And TG-SOF is validated as a single-pass decoder-boundary module and does not yet address multi-frame inputs, free-form generative VQA, or multi-step reasoning.
FGTR-Bench is audited and evidence-grounded, but its geographic, weather, and edge-case coverage remain incomplete, and higher benchmark accuracy should not be interpreted as a deployment safety guarantee under distribution shift.
Appendix~\ref{app:limitations} expands these points and discusses future work.

\newpage
\bibliography{ref}

\clearpage
\appendix

\section{FGTR-Bench Construction}
\label{app:dataset}

\noindent
This appendix collects FGTR-Bench construction material, model and training specifications, evaluation protocols, supplementary experimental results, limitations, and compliance notes referenced from the main paper.

\subsection{Data Sources}
\label{app:data_stats}
FGTR-Bench contains 40{,}236 single-image MCQ tuples for development: 34{,}749 training and 5{,}487 validation (86.4\% / 13.6\%). Each tuple pairs one image with one four-option question.
The five \emph{evaluation} tracks in the main paper---\textit{Holistic Sign}, \textit{Daytime Signal}, \textit{Nighttime Signal}, \textit{Roadside Micro}, and \textit{Participant Micro}---are reported on the blind test set described below.
We materialize split hygiene at the path level: every \texttt{image\_path} belongs to exactly one of train, validation, or test, with disjoint paths across all three so that neither validation nor test re-uses captures seen during adaptation.
The JSON \texttt{track} field stores stable upstream tags; headline numbers are obtained by a deterministic remap from \texttt{track} into the five published tracks, so ablations that need finer slices can be expressed as filters without regenerating questions.

\noindent\textit{Blind test set.}
In addition to the train/validation development corpus, we maintain a disjoint test set of 4{,}947 samples held out from both training and validation.
All FGTR-Bench accuracy numbers in the main paper (Tables~\ref{tab:main_results} and~\ref{tab:ablation_components}) are computed on this test split.
Its source breakdown is TT100K: 1{,}447, LISA: 1{,}500, and FGTR\_EXT (self-collected extension): 2{,}000.
Per-track test counts are Holistic Sign: 1{,}447; Daytime Signal: 750; Nighttime Signal: 750; Roadside Micro: 800; Participant Micro: 1{,}200.
The validation split (5{,}487 samples) is reserved for development and qualitative analysis only.

\noindent\textit{Upstream split.}
Source-level train/validation counts are: TT100K (3{,}899 / 1{,}447; total 5{,}346), LISA (19{,}350 / 2{,}349; total 21{,}699), and self-collected (11{,}500 / 1{,}691; total 13{,}191). The grand total is 40{,}236 pairs.

\begin{figure*}[!t]
    \centering
    \includegraphics[width=\textwidth]{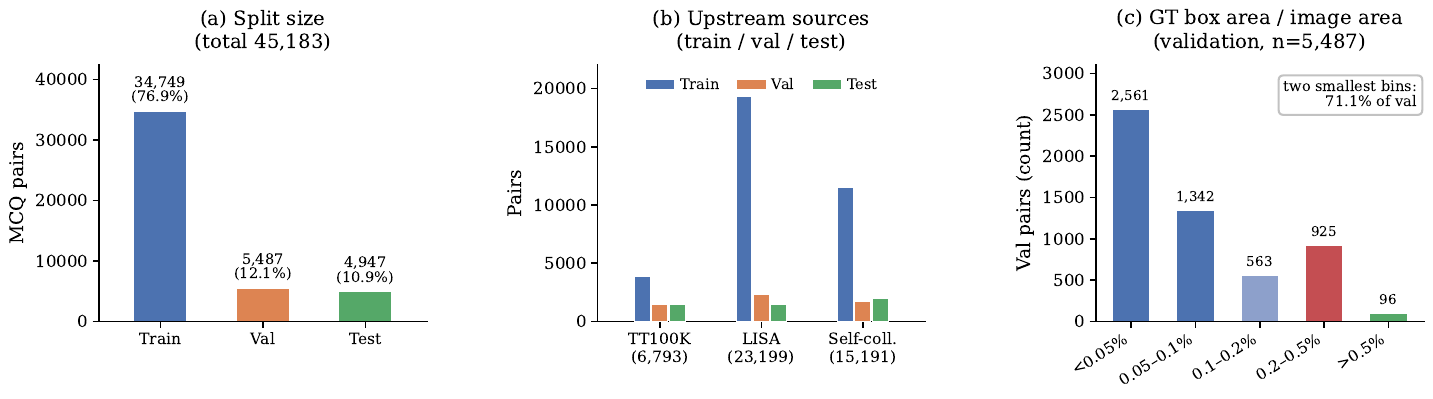}
    \caption{FGTR-Bench profile. \textbf{(a)} Train/validation/test MCQ counts (total 45{,}183; development corpus 40{,}236 is 86.4\%/13.6\% train/val). \textbf{(b)} TT100K, LISA, and self-collected pairs, each split into train, validation, and blind test (pair totals sum to 45{,}183). \textbf{(c)} Validation pairs binned by GT box-to-image area ratio using five bins (\(<\)0.05\%, 0.05--0.1\%, 0.1--0.2\%, 0.2--0.5\%, \(>\)0.5\%); counts sum to 5{,}487 and the two smallest bins account for 71.1\%.}
    \label{fig:app_benchmark_profile}
\end{figure*}

\subsection{QA Construction}

\paragraph{Export schema.}
Every released record follows the same machine-readable fields so training and evaluation scripts stay interchangeable.
Instead of a JSON listing, exports use the following keys:
\texttt{id} (unique string),
\texttt{split} (\texttt{train}, \texttt{val}, or \texttt{test}),
\texttt{image\_path} (relative path),
\texttt{bbox} (list of axis-aligned boxes or \texttt{null}),
\texttt{track} (unified taxonomy tag),
\texttt{source} (\texttt{tt100k}, \texttt{lisa}, or \texttt{self\_collected}),
\texttt{question} (stem),
\texttt{options} (four strings for \texttt{A}--\texttt{D}),
\texttt{answer} (gold letter),
and optional \texttt{meta} for provenance (\eg export version, generator ids).

\noindent\textit{Decoding contract.}
Models must emit a single choice letter; we score by exact match to \texttt{answer}.
Bounding boxes, when present, are image-axis-aligned and used for auxiliary alignment only.

\paragraph{Construction rules.}
TSR-Gen routes each annotated scene to a template family, samples hard negatives from \emph{same-scene confusers} where possible, and runs consistency checks (label--box--stem alignment) before export.
\smallskip
\noindent\textit{Sign-centric (TT100K).}
Stems tie to sign identity, lane context, or multi-sign composition.
Distractors are plausible confusers, not random open-vocabulary noise.
\smallskip
\noindent\textit{Signal-centric (LISA).}
Daytime vs.\ nighttime phrasing mirrors the atmosphere tag.
Distractors are other legal phases compatible with the same scene grammar.
\smallskip
\noindent\textit{Roadside / participant micro (self-collected).}
Questions anchor on small participants or hazards.
Negatives reuse nearby clutter categories.

We additionally enforce zero image-path overlap across train, validation, and test.

\paragraph{Annotators and auditors.}
\label{app:annotators}
Bounding-box annotation and MCQ auditing were carried out by four in-house contributors in two non-overlapping roles: two annotators labeled boxes and fine-grained tags, and two auditors reviewed exported MCQs before release.
All four are co-authors or direct participants in this project and were already familiar with the FGTR-Bench schema and evaluation goals, so no external recruitment or separate paid training was required.
The work was conducted solely for non-commercial academic research; participants received no salary or per-task payment beyond their regular research involvement.

\paragraph{Human annotation.}
\label{app:human_annotation}
Public sources are relabeled and self-collected frames are annotated in-house: the two annotators draw axis-aligned boxes, assign coarse/fine labels, and add lane or signal tags on safety-relevant entities, then link entities into a scene graph when spatial or relational cues matter.
These records are spot-checked before entering the frozen bundle below.

\paragraph{Ground-truth bundle (frozen inputs).}
\label{app:tsrgen_agents}
Every generation episode starts from a single image and a \emph{verified} structured record (human-audited or high-confidence auto-labels) that downstream agents may not contradict.
All agents read the same bundle: \texttt{image\_id}, \texttt{split}, \texttt{source}, \texttt{scene\_tags} (lighting and viewpoint cues such as day/night and ego/roadside), an \texttt{entities} array, and a \texttt{relations} array.
Each \texttt{entities} element carries \texttt{id}, axis-aligned \texttt{bbox}, \texttt{coarse\_class}, \texttt{fine\_label}, optional \texttt{lane\_id}, signal \texttt{phase} when applicable, and \texttt{attributes} (\eg occlusion).
Each \texttt{relations} element is a tuple $(e_s,e_o,\textit{type},\textit{detail})$; spatial links are typical.

\noindent\textit{Hard constraints.}
The correct MCQ letter must follow only from \texttt{entities} and \texttt{relations}.
Distractors come from labels present in the same \texttt{image\_id}, or from an approved confusion list for \texttt{fine\_label}.

\paragraph{Planner.}
The Planner reads the bundle and emits a \textbf{generation plan} that fixes which evidence entities are tested, which reasoning template applies (holistic sign, lane-linked sign, multi-sign, day/night signal, roadside micro, participant micro), how many distractors to draw, and which label pools are legal for negatives.
It does \emph{not} write natural-language options; it only fixes the discrete structure so later agents cannot drift from the boxes.
The Writer and Verifier consume a compact JSON plan with keys \texttt{template\_id}, \texttt{evidence\_entity\_ids}, an \texttt{answer} object (\texttt{letter}, \texttt{entity\_id}, \texttt{fine\_label} copied from the bundle), a \texttt{distractor\_policy} object (sampling pool, count $k$, and a \texttt{forbidden} list), a \texttt{question\_skeleton} string, and \texttt{difficulty} (\texttt{easy}/\texttt{medium}/\texttt{hard}).

\noindent\textit{System prompt (abridged).}
The following block is the instruction shell we attach to the Planner; concrete slot text is filled from the bundle.
\begin{quote}\footnotesize\ttfamily\raggedright
You are the PLANNER for FGTR-Bench\\
MCQ construction.\\
Input: JSON bundle (entities, relations, boxes).\\
Rules:\\
1) Pick ONE template\_id for the target track.\\
2) Pick evidence\_entity\_ids; union bbox\\
must support the gold answer only.\\
3) answer fields must copy the bundle;\\
never invent new labels.\\
4) distractor pool = same\_scene\_labels only.\\
5) Emit JSON only; no prose.
\end{quote}

\paragraph{Writer.}
The Writer receives the Planner JSON.
It expands \texttt{question\_skeleton} into a fluent stem and four options, aligning the gold letter with \texttt{answer.letter}.

\paragraph{Verifier.}
The Verifier checks letter--label agreement, distractor membership in the bundle (or approved confusion list), no contradictions with boxes, and a single best answer under the template.
Failed checks trigger a bounded replan or discard.

\paragraph{Manual audit protocol.}
\label{app:manual_audit}
After the Verifier and category filters, the two auditors manually audit a fixed slice of exported MCQs before release.
The audit focuses on whether each MCQ is answerable from visible evidence and whether the gold option is uniquely supported by the image.
For every audited record, auditors verify four items: (i) image--question relevance, (ii) answer uniqueness, (iii) distractor plausibility without creating label ties, and (iv) consistency between textual claims and annotated regions.

If any item fails, the sample is either corrected through bounded rewrite (stem/options only, without changing evidence entities) or removed from release.
Ambiguous cases are escalated to a second reviewer for adjudication.
This protocol is designed to suppress hallucinated image ties and near-duplicate option semantics that can inflate scores without improving evidence-grounded reasoning.

\subsection{Corpus Statistics}
Figure~\ref{fig:app_benchmark_profile} summarizes (i) train/validation/test scale, (ii) three-way upstream composition (TT100K, LISA, and self-collected streams) with train/val/test disaggregation, and (iii) a binned validation distribution of GT box area relative to image area.

Most decisive objects occupy a tiny image fraction, which motivates evidence-local MCQs rather than scene-prior shortcuts.
Panel \textbf{(c)} is therefore useful when diagnosing whether a model is rewarded for global scene priors: when the gold box occupies well below one percent of the image, lexical cues in the stem must still be resolved against a handful of high-frequency tokens rather than against generic context alone.
We keep the bin edges fixed so future dataset updates only swap counts, not definitions.

\begin{figure}[!t]
    \centering
    \includegraphics[width=\linewidth]{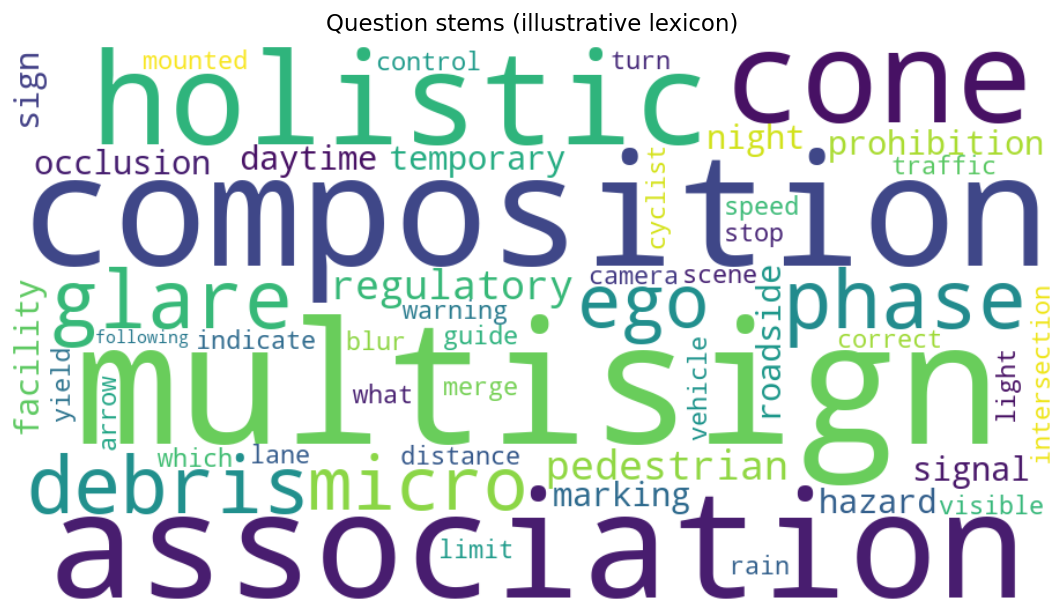}
    \vspace{0.45em}
    \includegraphics[width=\linewidth]{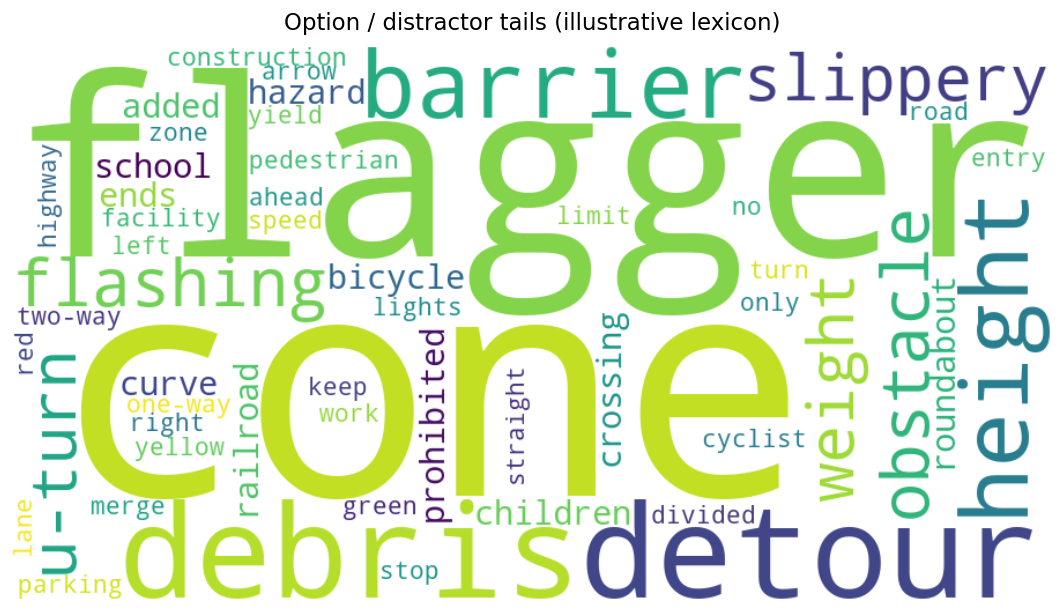}
    \caption{Corpus-based word clouds for question stems (top) and option/distractor vocabulary (bottom).}
    \label{fig:app_wordcloud_placeholder}
\end{figure}

For panel \textbf{(c)} in Figure~\ref{fig:app_benchmark_profile}, the validation-bin counts are \(<\)0.05\%: 2{,}561; 0.05--0.1\%: 1{,}342; 0.1--0.2\%: 563; 0.2--0.5\%: 925; \(>\)0.5\%: 96 (sum 5{,}487).

For \textbf{lexical EDA}, we report (a) length histograms of questions and full prompts, (b) option-letter balance at export time, and (c) word clouds over question stems and option/distractor tails generated from corpus token frequencies.
Figure~\ref{fig:app_wordcloud_placeholder} shows the finalized corpus-based clouds.

\section{Model Architecture Details}
\label{app:method}

This section supplements Secs.~\ref{sec:tg_sof}--\ref{sec:train_opt} with TG-SOF wiring and complexity details for reproduction on Qwen3-VL-4B.

\subsection{TG-SOF Module Details}
\label{app:tg_sof_impl}

\paragraph{Query-visible text.}
Salience uses only MCQ stems and the four option lines after templating.
We form $\mathcal{T}_{\mathrm{qv}}$ by dropping system prompts, image placeholder specials, and other non-task control tokens from $\mathcal{T}$.

\paragraph{Local detail prior ($\beta_j$).}
Vision tokens follow the raster order of the Qwen3-VL vision grid.
Let $\mathcal{N}(j)$ denote the one- or two-sided raster neighbors of token $j$.
We compute
\begin{equation}
\label{eq:app_beta_contrast}
c_j=\frac{1}{|\mathcal{N}(j)|}\sum_{j'\in\mathcal{N}(j)}\bigl\lVert \mathbf{h}^{(0)}_j-\mathbf{h}^{(0)}_{j'}\bigr\rVert_2,
\end{equation}
then min--max normalize $\{c_j\}_{j\in\mathcal{V}}$ within each sample to obtain $\beta_j\in[0,1]$ for Eq.~\eqref{eq:sharpen}.
The TG-SOF-only ablation row sets $\gamma{=}0$; the full TSR-MLLM learns $\gamma$ (Sec.~\ref{sec:tg_sof}).

\paragraph{Gate and residual heads.}
The gate MLP takes the concatenation of $\mathbf{h}^{(0)}_j$, mean-pooled query states over $\mathcal{T}_{\mathrm{qv}}$, $\tilde{\alpha}_j$ from Eq.~\eqref{eq:sharpen}, and $\beta_j$ (input dimension $2d{+}2$), uses two GELU layers with token- and sample-level sigmoid scaling, and outputs $\lambda_j$.
The residual head $\psi_\phi$ is a LayerNorm MLP with hidden width
\begin{equation}
\label{eq:app_psi_width}
d_{\psi}=\max(32,\lfloor 0.25d\rfloor).
\end{equation}
Trainable TG-SOF parameters are $\mathbf{W}_q$, $\mathbf{W}_k$, the gate MLP, $\psi_\phi$, and $\gamma$ when enabled.
We initialize $\psi_\phi$ and the gate path near zero so $\mathcal{F}_{\phi}$ starts close to identity on vision rows (Eq.~\eqref{eq:tg_sof_map}).
Top-$K$ selection and the vision update follow Eq.~\eqref{eq:tg_sof_update}; salience scoring uses Eq.~\eqref{eq:score_ij}--\eqref{eq:salience} with $\rho$ from Table~\ref{tab:app_train_recipe}.

\subsection{Complexity Analysis}
Per sample, bilinear scoring (Eq.~\eqref{eq:score_ij}) costs $\mathcal{O}(|\mathcal{T}_{\mathrm{qv}}|\,|\mathcal{V}|\,d_a)$, softmax normalization in Eq.~\eqref{eq:salience} costs $\mathcal{O}(|\mathcal{T}_{\mathrm{qv}}|\,|\mathcal{V}|)$, partial Top-$K$ selection contributes $\mathcal{O}(|\mathcal{V}|)$ when $K\ll|\mathcal{V}|$, and sparse residual application in Eq.~\eqref{eq:tg_sof_update} affects only $K$ rows at $\mathcal{O}(Kd)$.
These terms remain small relative to a full transformer forward through $\mathcal{G}_{\theta}$.
TG-SOF adds no extra image encoding, cropping, or fusion rewiring at inference.

\section{Training Details}
\label{app:train}

This section supplements Sec.~\ref{sec:train_opt} with the optimization recipe, training pipeline, and loss implementation used in our experiments.

\subsection{Optimization Recipe}
We fine-tune a stage-one adapter on Qwen3-VL-4B with the vision trunk and pretrained decoder blocks frozen; TG-SOF, decoder LoRA, and \texttt{lm\_head} are trainable.
Training uses the FGTR-Bench train split only ($34{,}749$ samples).
We optimize with AdamW (lr $5{\times}10^{-6}$, weight decay $0$), bf16 autocast, and gradient checkpointing, for one epoch ($1348$ optimizer steps) with per-device batch $1$, gradient accumulation $8$, and two data-parallel devices (effective batch $16$).
Top-$K$ gating keeps a fixed $5\%$ vision-token budget: every forward scores the full grid, but only the selected rows receive bounded residuals.

\paragraph{Low-rank adaptation (LoRA).}
\label{app:lora}
For each targeted linear layer we keep pretrained weights $\mathbf{W}_0\in\mathbb{R}^{d_{\mathrm{out}}\times d_{\mathrm{in}}}$ frozen and train a rank-$r$ residual
\begin{equation}
\label{eq:app_lora}
\mathbf{W}=\mathbf{W}_0+\frac{\alpha}{r}\,\mathbf{B}\mathbf{A},
\end{equation}
with $\mathbf{A}\in\mathbb{R}^{r\times d_{\mathrm{in}}}$ and $\mathbf{B}\in\mathbb{R}^{d_{\mathrm{out}}\times r}$.
LoRA is applied to every decoder self-attention projection ($q/k/v/o$) in all language layers; $\mathbf{A}$ is Gaussian-initialized, $\mathbf{B}$ starts at zero, and dropout $0.05$ is applied on the low-rank branch.
The vision encoder and feed-forward blocks remain without LoRA.

\paragraph{Per-step pipeline.}

Each training sample follows:
\begin{enumerate}
\item Encode $(\mathbf{x},q)$ to fused inputs $\mathbf{H}^{(0)}$.
\item Compute $(\alpha_j,\beta_j)$ via Eq.~\eqref{eq:score_ij}--\eqref{eq:salience} and Eq.~\eqref{eq:app_beta_contrast}--\eqref{eq:sharpen}, form the Top-$K$ mask, and apply Eq.~\eqref{eq:tg_sof_update} to obtain $\mathbf{H}^{(1)}$.
\item Run the decoder with LoRA active and evaluate Eq.~\eqref{eq:total_loss}.
\item Backpropagate through TG-SOF, LoRA, and trainable heads; optimizer step.
\end{enumerate}

\subsection{Loss Implementation}
\label{app:loss_impl}

\paragraph{Choice-token cross-entropy.}
$\mathcal{L}_{\text{ans}}$ is causal cross-entropy on supervised positions $\mathcal{Y}$---the gold MCQ letter plus required stop tokens---with prompts, image specials, and instruction prefixes masked.
This is the only objective used at inference.

\paragraph{Bounding-box alignment.}
When a record lists multiple boxes, we replace them by their pixel union $R$ before overlap testing.
Token $j$ is evidence-positive iff its Qwen3-VL vision patch intersects $R$:
\begin{equation}
\label{eq:app_patch_overlap}
a_j=\mathbb{1}\bigl[\text{patch}(j)\cap R\neq\emptyset\bigr].
\end{equation}
If $\mathcal{B}=\{j\in\mathcal{V}:a_j=1\}$ is empty, we skip $\mathcal{L}_{\text{bbox}}$ for that sample; otherwise the teacher is $q^{\text{bbox}}_j=a_j/|\mathcal{B}|$ as in Sec.~\ref{sec:train_opt}.
Salience logits, probabilities, and the KL term follow Eq.~\eqref{eq:salience_logit}--\eqref{eq:bbox_loss}; $\varepsilon$, $\tau$, and $\lambda_{\text{bb}}(t)$ are listed in Table~\ref{tab:app_train_recipe}.

\paragraph{Hidden-state consistency.}
The student forward enables TG-SOF and reads \emph{last-layer} decoder states at supervised positions $s\in\mathcal{S}$.
A detached teacher forward on the same inputs disables TG-SOF residuals and supplies teacher states at the same positions.
$\mathcal{L}_{\text{cons}}$ follows Eq.~\eqref{eq:cons_loss} and is applied only during training.

\paragraph{Ablation alignment.}
Under the same FGTR-Bench fine-tuning budget, cumulative ablation enables TG-SOF ($\gamma{=}0$), then the local detail prior ($\gamma$ learned), then $\mathcal{L}_{\text{bbox}}$, then $\mathcal{L}_{\text{cons}}$.

\subsection{Hyperparameter Details}
Table~\ref{tab:app_train_recipe} lists the full reported recipe.
\begin{table}[!t]
\centering
\small
\resizebox{\columnwidth}{!}{%
\setlength{\tabcolsep}{3.5pt}
\renewcommand{\arraystretch}{1.08}
\begin{tabular}{@{}ll@{}}
\toprule
\lcell{Setting} & \lcell{Value} \\
\midrule
\lcell{Base model} & \lcell{Qwen3-VL-4B; TG-SOF inserted\\before the decoder} \\
\lcell{Train data} & \lcell{FGTR-Bench training split only} \\
\lcell{Optimizer} & \lcell{AdamW, lr $5{\times}10^{-6}$,\\weight decay $0$} \\
\lcell{Schedule} & \lcell{$1$ epoch; $1348$ optimizer steps} \\
\lcell{Batching} & \lcell{per-device batch $1$; grad.\ accum.\ $8$;\\$2$ devices; effective batch $16$} \\
\lcell{Precision} & \lcell{bf16 training} \\
\lcell{Memory} & \lcell{gradient checkpointing enabled} \\
\lcell{Checkpoints} & \lcell{final adapter weights only} \\
\lcell{Throughput} & \lcell{$\approx$1.2 samples/s on our hardware\\($\approx$5.1\,h for one epoch)} \\
\midrule
\lcell{Vision packing} & \lcell{$12{,}544$--$12{,}845{,}056$ pixel bounds;\\resize on; square layout} \\
\midrule
\lcell{Supervision} & \lcell{causal cross-entropy on\\supervised choice tokens} \\
\lcell{LoRA} & \lcell{rank $64$, scaling $\alpha{=}128$, dropout $0.05$;\\all $q/k/v/o$ projections} \\
\lcell{TG-SOF adapter width} & \lcell{$\max(32,\lfloor 0.25\,h\rfloor)$\\with hidden size $h$} \\
\lcell{TG-SOF Top-$K$} & \lcell{keep ratio $\rho{=}0.05$;\\$K{=}\max(1,\lfloor0.05|\mathcal{V}|\rfloor)$} \\
\lcell{TG-SOF detail gain} & \lcell{learned $\gamma$; disabled ($\gamma{=}0$)\\in TG-SOF-only ablation row} \\
\lcell{TG-SOF gate} & \lcell{two-layer MLP (GELU) on\\$(2d{+}2)$-dim features; token/sample sigmoids} \\
\midrule
\lcell{Loss weights} & \lcell{$\lambda_{\text{bb}}(t)$ warmed from $0$;\\fixed $\lambda_{\text{cs}}$ in full model} \\
\lcell{BBox KL} & \lcell{temperature $\tau{=}1.0$;\\logit floor $\varepsilon{=}10^{-8}$} \\
\lcell{Consistency} & \lcell{teacher forward w/o TG-SOF;\\MSE on last-layer states at $\mathcal{S}$} \\
\bottomrule
\end{tabular}%
}
\caption{Fine-tuning recipe for the reported TSR-MLLM adapter.}
\label{tab:app_train_recipe}
\end{table}

FGTR-Bench tuples ship union boxes for auditing and KL sweeps; overlap tests follow the rasterization rules above.

\section{Evaluation Protocol Details}
\label{app:eval}

\subsection{FGTR-Bench MCQ Input}
Current online evaluation uses no system prompt (empty system field).
The user prompt keeps option letters \texttt{A}--\texttt{D} fixed and follows the template below.

\begin{quote}\footnotesize\ttfamily\raggedright
$\langle$question$\rangle$\\
A.\ $\langle$option\_1$\rangle$\\
B.\ $\langle$option\_2$\rangle$\\
C.\ $\langle$option\_3$\rangle$\\
D.\ $\langle$option\_4$\rangle$\\
Answer with the option's letter from the given choices directly.
\end{quote}

Models must emit a single choice letter; we score by exact match to \texttt{answer}.
Post-processing follows the online evaluator: \texttt{strip()} the generation; if the full string is one letter in \{\texttt{A},\texttt{B},\texttt{C},\texttt{D}\}, map directly; otherwise extract the first standalone capital letter via \verb|\b([A-Z])\b|; if still unmatched, normalize and try exact or substring match against options; unmatched outputs count as incorrect.
FGTR-Bench evaluation uses greedy decoding with a short answer cap ($16$ new tokens); bbox hints are disabled by default.
Main-paper accuracy is computed on the blind test set; the validation split is not used for those reported numbers.

\subsection{DriveQA-V Protocol}
\label{app:driveqa}

DriveQA-V (CARLA Signs) is our out-of-distribution transfer benchmark: models are adapted on FGTR-Bench only, then evaluated zero-shot on sign QA without any DriveQA supervision.
Table~\ref{tab:driveqa_signs_results} reports per-family accuracy on regulatory, warning, guide, and temporary control signs (17{,}375 MCQs in total), and Table~\ref{tab:ablation_components} repeats the overall DriveQA-V score for each cumulative ablation row.

We mirror the FGTR-Bench evaluation stack as closely as possible.
Images are packed with the same \texttt{min\_pixels} and \texttt{max\_pixels} bounds ($12{,}544$--$12{,}845{,}056$), resize-on square layout, and bf16 inference defaults as Appendix~\ref{app:train}.
Decoding is greedy with the same short letter cap ($16$ new tokens); bbox hints and task-specific prompts are disabled.
Predictions are scored with the official DriveQA-V letter-matching script so that baselines and TSR-MLLM remain comparable under a fixed post-processing contract.
When reporting the main-paper DriveQA-V row or the ablation transfer numbers, we do \emph{not} fine-tune on DriveQA; any FGTR-Bench-trained checkpoint is evaluated as-is on the full CARLA Signs split.

\section{Additional Experimental Results}
\label{app:additional}

\subsection{Baseline Details}
Table~\ref{tab:main_results} compares two FGTR-Bench adaptation regimes under a matched protocol.
Rows without FGTR-Bench training evaluate official checkpoints in a zero-shot setting on the 4{,}947-sample blind test set.

All supervised rows use the FGTR-Bench training split only (34{,}749 samples) and share the fine-tuning recipe in Table~\ref{tab:app_train_recipe}: frozen vision encoders, decoder LoRA (rank $64$, scaling $\alpha{=}128$, dropout $0.05$ on all $q/k/v/o$ projections), AdamW with learning rate $5{\times}10^{-6}$, weight decay $0$, bf16 training, effective batch $16$, and one training epoch.
Matched FGTR-adapted baselines (LLaVA-1.5, LLaVA-OneVision, Qwen2.5-VL-3B/7B, and Qwen3-VL-4B) differ only in backbone capacity; they omit TG-SOF and the auxiliary losses in Sec.~\ref{sec:train_opt}, and optimize answer-only cross-entropy.
TSR-MLLM uses the same backbone and LoRA schedule but adds TG-SOF plus the full objective in Eq.~\eqref{eq:total_loss}.

At inference, every reported model uses the same MCQ template, vision packing ($12{,}544$--$12{,}845{,}056$ pixels, resize on, square layout), greedy letter decoding, and a $16$-token answer cap (Appendix~\ref{app:eval}); bbox hints are disabled.

\subsection{Computational Cost and Efficiency}
We evaluate all main-table FGTR-Bench models on the full blind test set (4{,}947 MCQs) with batch size~$1$ per GPU under the protocol above.
A complete pass over the test split for our reported TSR-MLLM configuration takes approximately $1$\,h~$15$\,min on our hardware stack, which corresponds to $\approx$0.91\,s per image ($\approx$1.1\,images/s).

Vision encoding and the frozen decoder account for most of this wall time; TG-SOF adds one sparse Top-$K$ update (Eq.~\eqref{eq:tg_sof_update}) with negligible overhead relative to the backbone forward.

\subsection{Ablation Experiment}
We evaluate each cumulative configuration in Table~\ref{tab:ablation_components} with five random seeds and report FGTR-Bench blind-test overall accuracy per seed.
Paired differences between consecutive rows are tested with the Wilcoxon signed-rank test ($n{=}5$); we treat $p<0.01$ as significant.

Each added module yields a significant improvement over the previous row on FGTR-Bench: TG-SOF ($\gamma{=}0$) on the adapted Qwen3-VL-4B backbone ($\Delta{=}+0.8$\,pt, $p{=}0.0004$), the learned local detail prior ($\Delta{=}+0.5$\,pt, $p{=}0.0006$), the bounding-box focus loss ($\Delta{=}+0.4$\,pt, $p{=}0.0008$), and hidden-state consistency ($\Delta{=}+0.4$\,pt, $p{=}0.0010$), reaching 74.1\% overall for full TSR-MLLM.
All $p$-values are below the 0.01 threshold, indicating that the gains are unlikely to be seed noise.

The same cumulative stack improves DriveQA-V overall monotonically (71.7\%$\rightarrow$74.9\%) without any DriveQA fine-tuning; paired seed-level gains remain significant at each step ($p<0.01$), suggesting that the auxiliary objectives stabilize evidence use rather than overfitting FGTR-Bench alone.

\subsection{Additional Qualitative Results}
We supplement the main-text qualitative figure with failure cases that remain incorrect under TSR-MLLM.

\noindent\textbf{Success Case.}
Figure~\ref{fig:qualitative} in the main paper covers four validation tracks---\textit{Holistic Sign}, \textit{Signal Light}, \textit{Roadside Micro}, and \textit{Participant Micro}.
Each panel pairs the input frame with a TG-SOF importance overlay, the MCQ stem, and the verified answer ({\checkmark}), showing that sharpened query--vision salience $\tilde{\alpha}_j$ concentrates on the patch the question makes decisive (see \S\ref{sec:qualitative}).

\begin{figure}[!t]
    \centering
    \includegraphics[width=0.8\linewidth,trim=0.7cm 1.6cm 0.8cm 1.5cm,clip]{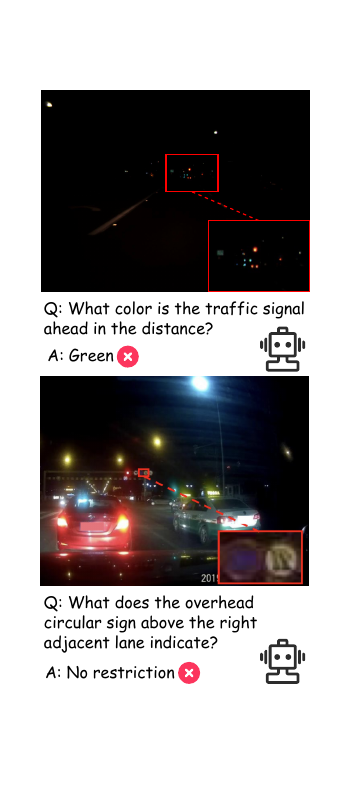}
    \caption{Representative TSR-MLLM failure cases under objective imaging degradations (incorrect predictions marked with {\color{red}\textbf{\texttimes}}).
    \textbf{Top:} low-light distant traffic signal; unresolved lamp color in the vision-token inputs leads to an incorrect \textit{Green} prediction on ``What color is the traffic signal ahead in the distance?''
    \textbf{Bottom:} nighttime motion blur on an overhead circular sign; smeared pictograms leave insufficient structure for TG-SOF to recover the correct restriction label, yielding an incorrect \textit{No restriction} answer.}
    \label{fig:app_fail_cases}
\end{figure}

\noindent\textbf{Failure Case Visualization and Analysis.}
Figure~\ref{fig:app_fail_cases} shows two remaining error modes driven by objective imaging degradations rather than ambiguous wording.

\noindent\textbf{Example 1 (low-light distant traffic signal).}
In Figure~\ref{fig:app_fail_cases} (\emph{top}), a distant signal at night occupies only a few vision tokens and its lamp color is unreliable after encoding.
TG-SOF can rerank existing slots but cannot reinstate contrast erased by low exposure, and the model incorrectly answers \textit{Green}; this pattern is common on \textit{Nighttime Signal} items.

\noindent\textbf{Example 2 (nighttime motion blur).}
In Figure~\ref{fig:app_fail_cases} (\emph{bottom}), streetlight glare, taillight bloom, and horizontal motion blur smear a boxed overhead circular sign in the inset crop.
High-frequency sign structure is largely lost before decoding, so the model incorrectly answers \textit{No restriction} despite a valid annotation box; this pattern is common on \textit{Holistic Sign} items captured at speed.

\section{Limitations and Future Work}
\label{app:limitations}
The main text (Section~\ref{sec:limitations}) summarizes the principal constraints; we detail them below and outline future directions.
While our results are promising, several limitations remain:
{\renewcommand{\labelenumi}{(\arabic{enumi})}
\setlength{\topsep}{0.35em}
\setlength{\itemsep}{0.2em}
\setlength{\parsep}{0pt}
\begin{enumerate}
\item Our study validates TG-SOF as a lightweight, query-conditioned module at the decoder boundary of a standard frozen MLLM backbone; extending it to multi-frame inputs or tighter region-level coupling is a natural next step while preserving single-pass inference.
\item FGTR-Bench is an audited single-image multiple-choice suite; scaling to broader geographic, weather, and edge-case coverage will require additional curation and domain-adaptive calibration.
\item The present framework has not yet been integrated with free-form generative VQA or multi-step reasoning pipelines supported by instruction-tuned MLLMs.
\item All experiments target traffic MCQ; beyond driving, TG-SOF should apply wherever decisive visual evidence is sparse relative to background (\eg medical imaging, remote sensing). We plan to validate the same query-guided, single-pass refinement recipe on additional fine-grained VQA domains to test cross-domain utility.
\end{enumerate}}
Combining query-guided token focus with generative decoding and chain-of-thought-style traffic reasoning is a promising direction for more general fine-grained scene understanding.
We leave these challenges to future work, aiming to advance reasoning systems that combine strong benchmark performance with reliable, evidence-grounded behavior across domains.

\section{Ethical and License Notes}
\label{app:ethics}
All public datasets and self-collected road images are used in compliance with licensing rules solely for non-commercial research.
Human annotation and MCQ auditing were performed in-house by project co-authors (Appendix~\ref{app:annotators}); no crowdsourcing platform was used and no per-task compensation was paid.

Self-collected frames, including those used for the \textit{Participant Micro} track, were reviewed for privacy-sensitive content before release.
Most participant instances are distant, low-occupancy targets in which individuals are not clearly identifiable; when faces, license plates, or other potentially identifying regions were visible at sufficient resolution, we applied blur-based de-identification before the images entered FGTR-Bench.
Public-source images are used under their original licenses; we do not redistribute any raw self-collected imagery beyond the released benchmark protocol.

FGTR-Bench is intended for research evaluation only and cannot guarantee practical deployment safety without formal verification, privacy review, and validation under distribution shift.

\section{Use of AI Tools}
\label{app:ai}
We used large language models for non-substantive writing assistance during manuscript preparation, including grammar polishing, LaTeX formatting, and literature organization.
LLMs also participate in the TSR-Gen QA construction pipeline (Appendix~\ref{app:dataset}); all exported MCQs pass automatic consistency checks and human audit before release.
All scientific claims, experimental configurations, reported numbers, and final wording were verified by the authors.

\end{document}